\documentclass{article}

\PassOptionsToPackage{numbers, compress}{natbib}

\usepackage[final]{neurips_2021}

\usepackage[utf8]{inputenc} 
\usepackage[T1]{fontenc}    
\usepackage{hyperref}       
\usepackage{url}            
\usepackage{booktabs}       
\usepackage{amsfonts}       
\usepackage{nicefrac}       
\usepackage{microtype}      
\usepackage{xcolor}         
	
\usepackage{graphicx} 
\usepackage{caption}        
\usepackage{setspace}       
\usepackage{array}   		
\usepackage{algorithm}
\usepackage[noend]{algpseudocode}	
\usepackage{wrapfig}

\newcommand{\stats}[3]{\hspace{-.25cm} \begin{minipage}{.55cm}{\ \\[-.4cm] \begin{spacing}{1}\begin{tabular}{>{\raggedleft}p{1cm}} \small #1 \\ #2 \\ #3 \end{tabular}\end{spacing}\vspace{.15cm}}\end{minipage}}
\newcommand{\alg}[1]{\begin{minipage}{1cm}{\small #1}\end{minipage}}

\newcolumntype{R}{>{\raggedleft}p{19.7mm}}

\newcommand{\ENC}{\mathsf{ENC}}
\newcommand{\AEnc}{\mathsf{AE}}
\newcommand{\proto}{\mathsf{proto}}

\newcommand{\bP}{\mathbf{P}}
\newcommand{\bz}{\mathbf{z}}
\newcommand{\by}{\mathbf{y}}
\newcommand{\bx}{\mathbf{x}}

\newcommand{\bg}{\mathbf{g}}

\newcommand{\calX}{\mathcal{X}}

\title{DeDUCE: Generating Counterfactual Explanations Efficiently}

\author{%
  Benedikt Höltgen\thanks{External collaborator to OATML Group, work done while a master student there.} \quad\quad Lisa Schut \quad\quad Jan M. Brauner \quad\quad Yarin Gal \\
  OATML Group \\
  Department of Computer Science\\
  University of Oxford\\
  Oxford, United Kingdom\\
  \texttt{benedikt.hoeltgen@mailbox.org}
}

\begin{document}

\maketitle

\begin{abstract}
    When an image classifier outputs a wrong class label, it can be helpful to see what changes in the image would lead to a correct classification.
    This is the aim of algorithms generating counterfactual explanations.
    However, there is no easily scalable method to generate such counterfactuals.
    We develop a new algorithm providing counterfactual explanations for large image classifiers trained with spectral normalisation at low computational cost.
    We empirically compare this algorithm against baselines from the literature; our novel algorithm consistently finds counterfactuals that are much closer to the original inputs.
    At the same time, the realism of these counterfactuals is comparable to the baselines. The code for all experiments is available at \url{ https://github.com/benedikthoeltgen/DeDUCE}.
\end{abstract}


\section{Introduction}

\begin{wrapfigure}{r}{.45\textwidth}
\centering
\vspace{-.6cm}
\includegraphics[width=\linewidth]{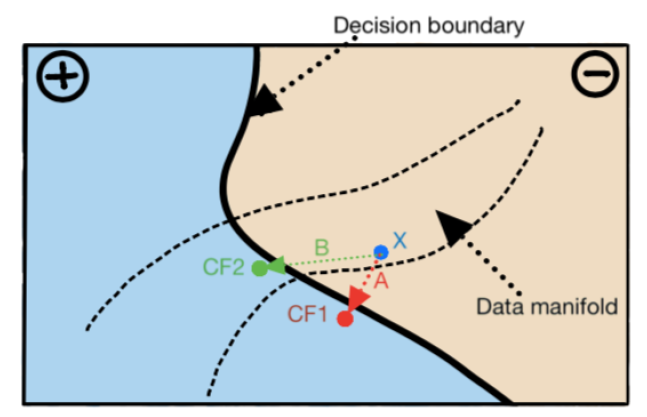}
\caption{In many use cases, counterfactuals should lie just across the decision boundary as seen from the original $x$ and lie on the data manifold, which is here satisfied by CF2 but not by CF1  (taken from \cite{Verma2020}).}
\label{fig:Verma}
\end{wrapfigure} 
Much of the recent staggering success of machine learning is due to large and complex models whose inner workings are in many cases elusive.
The prime examples of such 'black box' models are deep neural networks.
Despite the drawbacks that come with a poor understanding of their inner workings, such models are already widely deployed in practice due to their state-of-the-art performance in many areas.
But this poor understanding can make it difficult to interpret the decision making process of such models, and thus to identify the problem when a model makes an error.

To explain erroneous classifications, it can therefore be helpful to investigate the relevant decision boundary.
An intuitive way to do so is to look for alternative inputs that are similar but classified differently; 
this is the realm of counterfactual explanations.
For debugging, we usually want to generate realistic counterfactuals that are just across the model's classification decision boundary (fig.~\ref{fig:Verma}).
We want a counterfactual that stays as close to the original input as possible but results in a correct classification, in order to understand the system's decision making.
At the same time, the counterfactual should stay on the manifold of realistic images (CF2 in fig.~\ref{fig:Verma}) since we are interested in inputs that can be interpreted.
Being a nascent area of research, there is so far no algorithm providing such counterfactual explanations for large models at low cost.
In this work, we introduce a fast novel algorithm using specific properties of residual networks (ResNets), a widely used model, to find counterfactuals which are very close to the original input.

The contributions presented in this work are the following:
\begin{itemize}
\item We introduce and implement a new, scalable algorithm for generating counterfactual explanations in residual networks trained with spectral normalisation. In comparison to previous approaches, this algorithm has a low engineering overhead as well as low computational requirements; it does not rely on auxiliary generative models and can run on single forward pass residual networks rather than on ensembles of models.
\item We propose and implement a way to assess realism in counterfactual explanations, drawing on the literature on anomaly detection.
\item We evaluate the novel algorithm on the MNIST dataset and compare its performance against three baselines from the literature.
\end{itemize}


\section{Background}

\subsection{Counterfactual explanations}
\label{ssec:CEs}

Counterfactual explanations for machine learning models have been introduced in \citep{wachter2017}.\footnote{A similar notion of counterfactuals in ML has also been explored by \citep{Kusner2017} in the context of algorithmic fairness rather than explainability.}
They are used as post-hoc explanations for individual decisions.
In general, a counterfactual explanation is understood as the presentation of an alternative input $\bx'$, the counterfactual, which is in some way similar to the original $\bx$ yet leads to a different prediction.
Due to their similarity (both conceptually and w.r.t. the algorithms that generate them), counterfactuals are often introduced in juxtaposition with adversarial examples (AEs).\footnote{This is only partly helpful since there is also no consensus on the definition of AEs. While scholars generally agree that AEs are necessarily misclassified, there is no consensus on whether they need to be generated by imperceptible perturbations. Another, less operational definition could be that they are not detected to be outside the distribution of realistic datapoints -- although the notion of `misclassification' could already be interpreted as requiring the datapoint to be in-distribution as it implies that there is a correct classification.}
\citet{Freiesleben2021} provide an illuminating discussion of various definitions in the literature and come to the conclusion that while AEs are necessarily misclassified, counterfactuals need not be (and often should not be).
Furthermore, in agreement with \citep{Molnar2020}, they define counterfactuals as the \emph{closest} alternative input (on some suitable metric) that changes the prediction (to a pre-defined target, if applicable).
This seems to be too restrictive in general as there are applications that do not require such a strong focus on proximity.
Furthermore, the definition then depends too much on the choice of the similarity metric.
Consequently, we will generally call an alternative input $\bx’$ a counterfactual to an input $\bx$ under model $f$ if it is similar to $\bx$ on a suitable similarity relation but changes the prediction of the model.\footnote{One might drop the requirement of changing the prediction for applications beyond explanations, e.g. for assessing fairness \citep{Kusner2017}.}

Counterfactual explanations can be useful for debugging:
They can be used to answer questions like `Why did the self-driving car misidentify the fire hydrant as a stop sign?' \citep{Goyal2019}.
More precisely, they can answer the questions 'What would the image need to look like in order to be classified as a fire hydrant?' and 'What changes would need to happen in other fire hydrant (stop sign) images in order to be classified as a stop sign (fire hydrant)?'.
Depending on the model and the application, we might only be interested in counterfactuals that look realistic.
In particular, we learn more about the decision boundary when we understand the changes made to the image.
If, on the other hand, the classification is changed by adding unrealistic noise, then we only learn that the classifier is not robust to this.
In terms of \citet{lipton2018}, such counterfactuals are less informative.
Counterfactual explanations are often required to be realistic \citep{schut2021}, likely \citep{Molnar2020}, or plausible \citep{Karimi2021}.
Perhaps the most useful way to formulate it is to require counterfactuals to be `likely states in the (empirical) distribution of features' which are `close to the data manifold' \citep{Karimi2021}.
Another often-mentioned desideratum for counterfactual explanations is sparsity: in general, the less features are changed in input space, the better.
Sparse perturbations are usually more interpretable as the change in classification can be attributed to a smaller part of the input which is easier to grasp both in terms of the model behaviour and in terms of the input itself.
If all parts of the input change a little, it might be harder to understand what the perturbation means and how it affects the model.

\subsection{Epistemic uncertainty}
\label{ssec:EU}

Previous work has shown that epistemic uncertainty can be used as a proxy for realism \cite{smith2018, schut2021}.
There are two kinds of uncertainty relevant to machine learning models.
Aleatoric uncertainty, on the one hand, is inherent in the data distribution and cannot be reduced. 
It is high when there is no clear ground-truth label and maximised in the extreme case of random labels.
Epistemic uncertainty, on the other hand, is due to a lack of knowledge on the part of the model.
It is a quantity that can be reduced for a given input  by including it in the training set.
Given this characterisation, epistemic uncertainty estimates are used for active learning (selecting samples that are particularly useful to train on) as well as for detecting out-of-distribution inputs.

It has also been observed that neuron activations in late layers of a neural network, which we call \emph{features}, can be used to estimate epistemic uncertainty when two properties called sensitivity and smoothness are satisfied by the mapping into the feature space \cite{VanAmersfoort2020}.
Sensitivity can be seen as a lower Lipschitz bound, ensuring that the features remain sensitive to differences between inputs, thereby preventing `feature collapse':
Otherwise, out-of-distribution (OoD) inputs might not be distinguishable from in-distribution (iD) inputs as they could be mapped to the same area in feature space.
Conversely, smoothness can be seen as an upper Lipschitz bound: 
Similar inputs are guaranteed not to be too far from each other in feature space, such that distances in this space remain meaningful.
Building on \cite{Bartlett2018}, \citet{Liu2020} show that applying spectral normalisation (SN, \cite{miyato2018}) with a coefficient $c \leq 1$ to ResNets \cite{he2016} is enough to enforce both sensitivity and smoothness.
Using such models, \citet{mukhoti2021} fit a probability distribution to the feature space after the last ResNet block, using the feature representations of the training data.
To estimate the epistemic uncertainty of an input, they then calculate the negative log-likelihood of its feature representation under the learned distribution.
For the probability distribution, they use a Gaussian mixture model (GMM).
We will make use of this approach, called Deep Deterministic Uncertainty (DDU), in the algorithm we propose below.

\subsection{Related work}
\label{ssec:related}

In this section, we provide a (non-exhaustive) survey of algorithms that provide counterfactual explanations.
In the initial paper, \citet{wachter2017} propose to simply optimise the objective 
\begin{equation}
\arg \min_{\bx'} \max_\lambda\ (f(\bx') - \by')^2 \cdot \lambda + d(\bx, \bx')
\end{equation}
where $f$ denotes the model, $\by'$ the desired model prediction, $d$ a pre-defined distance metric, and $\lambda$ is a hyper-parameter.
They suggest to iterate through increasing values of $\lambda$, always solving for $\bx'$ for fixed $\lambda$, until a counterfactual sufficiently close to the original input is found. 
This is quite a minimal approach, which comes at the cost of $\bx'$ not being constrained to lie on the data manifold (which might be required, depending on the task).

\citet{VanLooveren2020} build on this (and on \cite{Dhurandhar2018}), but focus on generating more interpretable counterfactuals by optimising a more complex objective function.
Their prototype-guided approach minimises the loss
\begin{equation}
c L_{pred} + \beta L_1 + L_2 + L_{AE} + L_{proto}.
\end{equation}
Here, $L_{pred} = f(\bx')_l - \max_{i \neq l} f(\bx')_i$ where $f(\bx')_l$ is the softmax output of the classifier $f$ on the counterfactual $\bx'$ for the original class $l$, i.e. its confidence that $\bx'$ belongs to class $l$.
$L_1$ and $L_2$ refer to the corresponding distances between $\bx$ and $\bx'$ in input space.
$L_{AE} = \gamma \|\bx' - \AEnc(\bx')\|_2^2$ where $\AEnc$ is an autoencoder trained on the training data.
Lastly, \mbox{$L_{proto} = \theta \| \ENC(\bx') - \proto_t \|_2^2$} where $\proto_t = \frac{1}{K} \sum_{k=1}^K \ENC(\bx_k^t)$ is the latent prototype defined by the $K$ nearest neighbours $\bx_k^t$ of $\bx'$ in target class $t$ and $\ENC$ is an encoder.
For untargeted counterfactuals, $t$ is chosen as $\arg\min_{t \neq l}\| \ENC(\bx') - \proto_t \|_2$.
While $L_{pred}$ enforces a change of classification, $L_{AE}$ and $L_{proto}$ are included to encourage realism, measured by a low reconstruction loss under the autoencoder and similarity to similar training samples in the encoded latent space.
The additional use of an autoencoder and the additional loss terms generate a computational overhead that slows down the approach.
Furthermore, the approach comes with many hyperparameters which might require a lot of tuning when applying it to a new task.
The interplay of the different loss terms is not straightforward to analyse either formally or conceptually, so it is hard to predict and improve the performance on new applications.

A very different route to provide interpretable counterfactuals is taken by \citet{schut2021}, based on the notion of uncertainty \citep{gal2016}.
They suggest that counterfactuals are realistic if the model classifies them with low epistemic uncertainty and unambiguous when the model has low aleatoric uncertainty.
Consequently, they propose to minimise overall uncertainty in models that provide accurate uncertainty estimates through their softmax output, such as deep ensembles \citep{Lakshminarayanan2017}.
They show that maximising the ensemble's target class prediction $f(\bx')_t$ is sufficient to minimise its predictive entropy and hence both epistemic and aleatoric uncertainty.
Rather than using an off-the-shelf optimisation algorithm as the two previously mentioned approaches, Schut and colleagues compute the gradient of the classification loss in input space and identify the most salient pixel for reducing this loss.
Similar to JSMA \citep{Papernot2016}, they iteratively change the most salient pixel until the target prediction is above 99\%, when the uncertainty is sufficiently low.
This works well in practice but using an ensemble of models (for their MNIST experiments they use 50 models) is computationally expensive, which might hinder its deployment in practice.

Another approach that has been suggested both for providing counterfactual explanations and for algorithmic recourse is REVISE \citep{Joshi2019}.
This algorithm requires the availability of a generative model, such as the decoder of a VAE trained on the training data.
Similar to \citep{wachter2017}, the overall idea is to minimise the function
\begin{equation}
\ell(f(G(\bz')), t) + \lambda \cdot \| G(\bz') - \bx \|_1
\end{equation}
where $f$ is the classifier, $t$ is the target, $\ell$ is some loss function, and $G$ is the generative model.
To find a $\bz'$ that minimises the loss, $\bz$ is initialised to the encoding of the original input $\bx$; then the gradient of the loss in the latent space is computed and the algorithm iteratively takes small steps in latent space until the prediction changes to the target.
Since the resulting counterfactual $\bx' = G(\bz')$ is produced by the generative model, it can be dissimilar to $\bx$:
Although the $L_1$ norm is known to encourage sparsity, the algorithm cannot be expected to provide sparse solutions, as the changes are not taken in the input space.
Another disadvantage of using a generative model is, of course, the need to train it beforehand which can pose difficulties of varying degree depending on the data domain.
Several recent works utilise GANs for generating counterfactuals, such as \cite{kenny2021}.


\section{Novel algorithm: DeDUCE}
\label{sec:deduce}

\begin{wrapfigure}{r}{.45\textwidth}
\centering
\vspace{-.5cm}
\includegraphics[width=\linewidth, trim={1.7cm 1cm 1cm 1cm},clip]{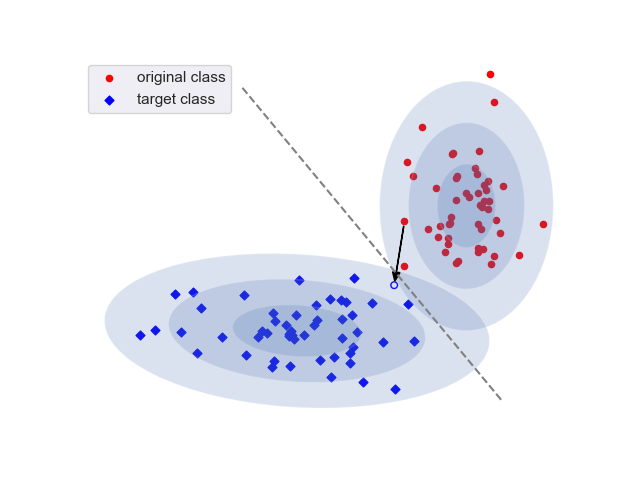}
\caption{The training data is used to fit a Gaussian distribution for each class in a (here depicted) feature space. DeDUCE then changes a given input in a way that increases the density under the target class Gaussian in the feature space, until the decision boundary (dashed line) is crossed. This allows to generate realistic counterfactuals.}
\label{fig:new_fig1}
\end{wrapfigure} 
For large image datasets, ResNets are often the model of choice.
When ResNets are trained with spectral normalisation, we can use DDU (section~\ref{ssec:EU}) to estimate their epistemic uncertainty.
DDU achieves state-of-the-art results in OoD detection such as MNIST vs. FashionMNIST.
The authors also demonstrate that it can be used for active learning;
this implies that it is particularly useful to train on inputs whose representations have low likelihood under the GMM, i.e. such inputs are substantially different from the previous training data.
This suggests that DDU's measure of epistemic uncertainty can be a useful target when aiming to generate counterfactuals that are similar to the original training data.
The idea of the novel algorithm presented here is to cross the model’s decision boundary while keeping the epistemic uncertainty as low as possible, using DDU.
Instead of maximising the whole GMM density for minimising epistemic uncertainty (i.e. maximising feature-space density), we propose to only maximise the target class density (fig.~\ref{fig:new_fig1}).
Otherwise, one would also maximise the original label's class density, which could lead to unstable behaviour.
As we also want to cross the decision boundary quickly, we suggest to  change the pixels that are most salient for the gradient of the loss
\begin{equation}\label{eq:loss_fct}
\bg = \nabla_\bx \left( \ell_c(\bx, t) \cdot \lambda - \log p_t(f_Z(\bx)) \right)
\end{equation}
where $f$ is the model, $t$ the target class, $\ell_c$ the cross-entropy, $p_t$ the target class density, and $f_Z$ the feature extractor, i.e. the part of the model that maps to the feature space after the last ResNet block.
This means we are trying to change the input in a way that quickly changes the classification (first term) and makes it more similar to the target class in feature space (second term).
The first term typically has values in $[0, 100]$, while the second term can have values as low as $10^{-6}$.
Instead of changing the pixels that minimise the weighted loss for some $\lambda$, we select the pixels which make the largest relative difference to either of the two loss terms.
Therefore, we use the alternative gradient
\begin{equation}\label{eq:grad_sum}
\bg = \frac{1}{\ell_c(f(\bx'), t)} \nabla_{\bx'} \ell_c(f(\bx'), t) \cdot \mu - \frac{1}{| \log p_t(f_Z(\bx')) |} \nabla_{\bx'} \log p_t(f_Z(\bx'))
\end{equation}
instead of the gradient of the weighted loss given in equation~(\ref{eq:loss_fct}).
Note that the cross-entropy loss $\ell_c(f(\bx'), t)$ is non-negative whereas $\log p_t(f_Z(\bx'))$ can be negative or positive, depending on whether the density is above or below one.
Despite their similarity, including both terms in the objective indeed improves the quality of generated counterfactuals; we also find that using the alternative gradient without further weighting ($\mu=1$) works better than the gradient of the loss for any value of $\lambda$ (see appendix~\ref{app:gradients}).

The novel approach that we call \mbox{DeDUCE} (Deep Deterministic Uncertainty-based Counterfactual Explanations) is described in algorithm~\ref{alg}:
In order to only make small and sparse changes to the original input, DeDUCE iteratively perturbs only few pixels at a time.
At each iteration, it determines the most salient pixels for maximising the objective, by computing the gradient in input space, and then perturbs them by a fixed step size $\delta$.
This is similar to how the Jacobian-based Saliency Map Attack (JSMA) \citep{Papernot2016} generates adversarial examples and the approach of \citep{schut2021} generates counterfactuals.
Thereby, DeDUCE iteratively perturbs the input in small steps in a way that makes it more and more similar to the target class until it crosses the decision boundary.
The algorithm stops when the softmax output for the target class is above 50\% as this corresponds to the model choosing `in target class' over `not in target class'.
Following \citep{schut2021}, DeDUCE limits the number of times each pixel can be updated\footnote{This is achieved by counting the number of updates per pixel in $\bP$ and applying a mask to the gradient $\bg$ that sets $\bg[i]$ to 0 if $\bP[i] \geq r$. The expression ${\tt select\_q\_largest\_masked}(|\bg|,\ \bP < r)$ in line~\ref{alg:select} denotes the procedure of first applying this mask and then selecting the positions of the $q$ largest values of $|\bg|$.\label{fn:pixels}} and clips $\bx'$ to the input domain bounds.
Similar to some work on adversarial examples \cite{dong2018}, we also add momentum to the gradient, replacing the expression for $\bg$ by 
\begin{equation}
\bg_k = \frac{\nabla_{\bx_k} \ell_c(f(\bx_k), t)}{\ell_c(f(\bx_k), t)} - \frac{\nabla_{\bx_k} \log p_t(f_Z(\bx_k))}{| \log p_t(f_Z(\bx_k)) |} + m \cdot \bg_{k-1},
\end{equation}
with $\bx_k$ referring to the state of the input after $k > 0$ iterations.
For the experiments reported below, we change one pixel at a time and use a momentum of 0.6.
Adding momentum of this size often does not make a difference; in our experiments, it only affected around 1.3\% of the generated counterfactuals.

\begin{algorithm}[h]
\caption{DeDUCE}
\textbf{Inputs:} original input $\bx \in \calX$, target class $t$, trained model $f$ with feature extractor $f_Z$, training data $X_{tr} \in \calX^N$, coefficient $\lambda$, step size $\delta$, max iterations $\mathsf{max\_iter}$, max pixel changes $p$, number of pixels $m$, target confidence $\gamma$.\\
\textbf{Output:} counterfactual $\bx' \in \calX$
\begin{algorithmic}[1]
\State (before deployment) apply $f$ to $X_{tr}$ and  fit the GMM $p(\bz) = \frac{1}{|C|} \sum_{c \in C} p_c(\bz)$
\State $\bx' \leftarrow \bx$
\State $k \leftarrow 0$
\State $\bP \leftarrow \mathbf{0}_{dim(\calX)}$
\While {$f(\bx')_t \leq \gamma\ \mathbf{and}\ k < \mathsf{max\_iter}$}
\State compute gradient $\bg$ in input space
\State select $m$ most salient pixels: $I = {\tt select\_q\_largest\_masked}(|\bg|,\ \bP < p)$ \label{alg:select}
\State update these pixels: $\forall i \in I: \bx'[i] \leftarrow \bx'[i] + {\tt sign}(\bg[i]) \cdot \delta$
\State clip to input domain: $\bx' \leftarrow {\tt clip}(\bx')$
\State $\forall i \in I: \bP[i] \leftarrow \bP[i] + 1$
\State $k \leftarrow k+1$ 
\EndWhile
\State \textbf{return} $\bx'$
\end{algorithmic}\label{alg}
\end{algorithm}


\section{Experiments}


\subsection{Dataset and metrics}

We perform experiments on the MNIST dataset \cite{lecun1998}, so far the only widely used image dataset in the literature on counterfactual explanations.
We use the $L_1$ and $L_0$ distances in input space to assess similarity and sparsity, respectively.
We also want to measure how realistic the generated counterfactuals are, as this is usually helpful (cf.~section~\ref{ssec:CEs}).
There is no consensus in the field on how to measure realism (or 'plausibility' \cite{keane2021}), and for images, this proves to be quite difficult.
Therefore, we turn to the literature on anomaly detection and use the approach that performed best on an anomaly detection task for MNIST in a recent study \cite{ruff2021}.
This approach, called AnoGAN \cite{schlegl2017}, uses a pre-trained generative adversarial network (GAN, \cite{goodfellow2014}) to compare the similarity of a given input $\bx$ with the closest image  $G(\bz)$ that the GAN can generate. 
AnoGAN uses gradient descent in latent space to minimise the loss
\begin{equation}\label{eq:agan_obj}
\| G(\bz) - \bx \|_1 + \lambda \cdot \| f_D(G(\bz)) - f_D(\bx) \|_1,
\end{equation}
where $G$ is the generator and $f_D$ is a mapping to a later layer of the discriminator.
The first term gives the $L_1$ distance between the generated sample and the input whereas the second term measures how similar their feature representations are in the discriminator model.
We use a Wasserstein GAN \cite{arjovsky2017} trained on MNIST.
To reduce the dependence on the initial $\bz$, we perform gradient descent three times from different, randomly sampled starting points.
We use $\lambda = 0.1$, an initial learning rate of $0.03$, and bound the number of iterations to 4000.
We tested how well the resulting metric allows to distinguish actual MNIST images from EMNIST character as well as Fashion-MNIST images.
Our AnoGAN method achieves an AUROC of 0.913 and 0.998, respectively.
Note that these comparisons are quite different to the generated counterfactuals and thus only provide general sanity checks rather than actual performance tests.

\subsection{Baselines}
\label{ssec:baselines}

To assess the performance of the novel algorithm, we compare it with three baselines applied to ResNets without spectral normalisation.
The prototype-guided approach that we shall call `VLK' \citep{VanLooveren2020} as well as REVISE \citep{Joshi2019} were discussed in section~\ref{ssec:related}.
Their selection is largely based on a general scarcity of algorithms that provide counterfactual explanations, were demonstrated on image data, and are applicable to ResNets.
Although also demonstrated on MNIST, the mentioned approach of \citep{schut2021} is not included since it cannot be applied to single ResNets.
In order to get the required calibrated uncertainty outputs, one would need an ensemble of around ten models \citep{Lakshminarayanan2017}, which would make the results much less comparable.
In addition to VLK and REVISE, we include JSMA \citep{Papernot2016} as a third baseline.

It should be noted that \textbf{JSMA} was introduced to generate (perceptible) adversarial examples rather than counterfactual explanations, so the comparison with regard to realism is not a fair one.
However, since DeDUCE is loosely based on JSMA, the comparison is interesting as it shows how the modifications affect the results.
While the original paper recommends changing two pixels at a time, we change one as this makes it perform better in our setting and more comparable to the used DeDUCE algorithm.

To generate counterfactuals with \textbf{VLK}, we use the authors' implementation in the {\tt alibi} package \citep{alibi} in order to be as faithful as possible.
As the algorithm is already tuned to MNIST, we only make one change to the default setting, namely setting $k$ for $k$-nearest (encoded) neighbours in the $L_{proto}$ term to 5.
This is recommended in the paper and generally improves the quality of the generated counterfactuals.
Note that VLK uses an optimisation algorithm that includes the model's target confidence in the $L_{pred}$ loss term.
This means that, contrary to the other three algorithms, we cannot prescribe the generated counterfactuals to have a target confidence just above 50\%.
In fact, they have a mean confidence of 0.42 and standard deviation of 0.44, with many values being close to 0 or 1.

The third baseline \textbf{REVISE} requires more tuning, as it has not been demonstrated on MNIST before.
REVISE is designed to be applicable to image data, with the authors providing a demonstration on the CelebA dataset.
Sample reconstructions of the used VAE are shown in appendix~\ref{app:revise}.
Note that a more powerful generative model than the one employed here could improve the quality of the generated counterfactuals; this might, however, come with even higher engineering efforts and computational costs.
Appendix~\ref{app:revise} provides more details on the implementation of REVISE.

\subsection{Results}
\label{ssec:results}

With the four algorithms tuned to the dataset and base model, we can look at their performance.
We use five sets of 100 original MNIST images from the test set and try to find counterfactuals for all 9 potential target classes, resulting in $5 \times 900$ runs overall for each algorithm.
The quantitative metrics are the $L_0$ and $L_1$ distance to the original, and the rate at which the algorithms fails to output a candidate counterfactual.
For the evaluation of how realistic/anomalous the images are, we use the anomaly detection metric `AGAN' using AnoGAN.
The results are reported in table~\ref{tab:results}.
In order to ensure a fair comparison, only image-target pairs are included for which all algorithms found a counterfactual.

\begin{table}[h]
    \caption{Overall means from 5 $\times$ 900 image-target pairs, with standard deviations over the five means in brackets. \emph{AGAN} denotes the AnoGAN score. Original images achieve an average \emph{AGAN} score of $15.44$. Lower is better everywhere.\\} \label{tab:results}
    \centering
    \begin{tabular}{lrrrr}
        \toprule \centering
        algorithm & \multicolumn{1}{c}{\emph{AGAN}} & \multicolumn{1}{c}{$L_0$} & \multicolumn{1}{c}{$L_1$} & \multicolumn{1}{c}{\emph{failure in \%}} \\
        \midrule
        DeDUCE & 22.51 (0.72) & \textbf{21.16} (0.46) &  \textbf{10.72} (0.10) & \textbf{0.00} (0.00) \\
        JSMA & 23.90 (0.41) & 25.65 (0.65) &  12.63 (0.27) & 3.09 (0.28)\\
        VLK & 22.95 (0.93) & 155.43 (4.15) &  38.95 (1.19) & 0.09 (0.20)\\
        REVISE & \textbf{20.09} (0.88) & 752.46 (4.98) &  53.86 (0.58) & 26.80 (1.03) \\
        \bottomrule
    \end{tabular}
\end{table}

REVISE achieves the best results on the AnoGAN metric, but their scores are not as good as for the original images, which get a score of $15.44 \pm 2.04$.
VLK and DeDUCE do not show significant differences on this metric. 
We note that the directly reconstructed images under the REVISE VAE (without applying REVISE) are judged to be more realistic than the original images, achieving a score of $14.64 \pm 0.23$.
This shows that the metric has a bias towards VAE-generated images, which benefits REVISE and perhaps also VLK, as the latter minimises an autoencoder loss.
DeDUCE achieves better results than JSMA on all four metrics, with these differences are all being highly significant ($p < 10^{-7}$ on paired t-tests).
Both VLK and REVISE perform much worse than the other two on the sparsity as well as the similarity metric.
This is also expected since both DeDUCE and JSMA take pixel-wise steps in the input space.
The standard deviations on all metrics are generally fairly low, especially for DeDUCE and JSMA.

There are great differences with respect to the required computation times, not only because the provided implementation of VLK does not support the generation of counterfactuals in batches.
Even when all approaches are used to generate counterfactuals individually, VLK is the slowest approach (table~\ref{tab:times}).
JSMA is slightly faster than DeDUCE, which is still significantly faster than REVISE.
REVISE has a large standard deviation because some runs take particularly long.
Note that REVISE could work with a slightly larger step size and with a lower number of iterations, at the expense of quality.
All computations were performed on NVIDIA Tesla T4 GPUs.
\begin{center}
    \captionof{table}{Average running times (STDs in brackets) for generating counterfactuals individually.} ~\\
    \label{tab:times}
    \begin{tabular}{lr}
        \toprule \centering
        algorithm & \multicolumn{1}{c}{\emph{time in sec}} \\
        \midrule
        DeDUCE & 2.99\ \ \ (1.69) \\
        JSMA & \textbf{1.01}\ \ \ (0.52)\\
        VLK & 109.86\ \ \ (1.54)\\
        REVISE & 46.66 (84.33) \\
        \bottomrule
    \end{tabular}
\end{center}

\begin{figure}
\begin{minipage}{.06\linewidth}
\
\end{minipage}
\begin{minipage}{.13\linewidth}
\raggedright
\def\yskip{1.28cm}
original\\[\yskip]
DeDUCE\\[\yskip]
JSMA\\[\yskip]
VLK\\[\yskip]
REVISE
\end{minipage}
\begin{minipage}{.8\linewidth}
\begin{flushleft}
\includegraphics[width=.9\linewidth]{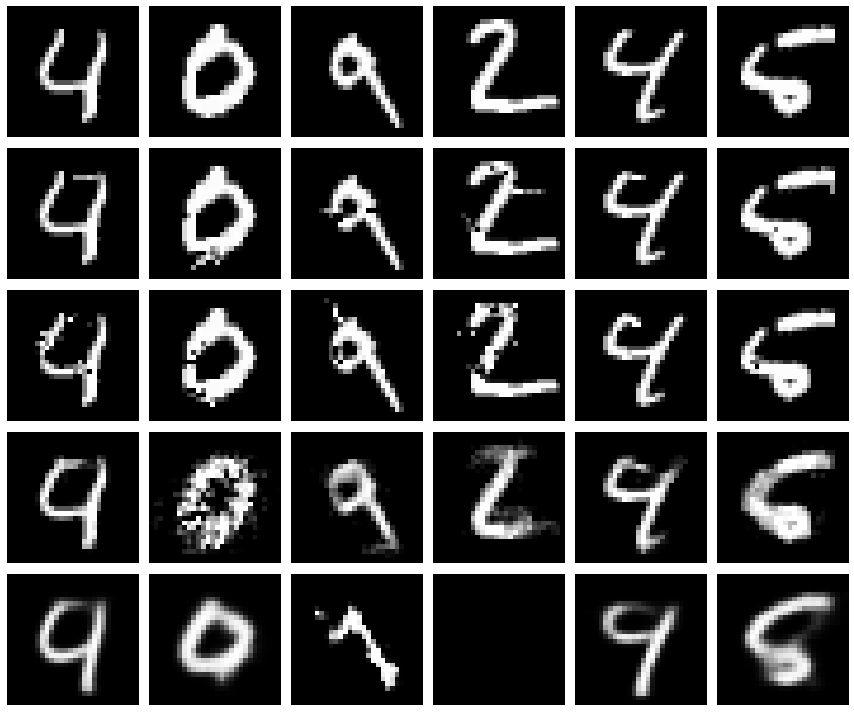}
\end{flushleft}
\end{minipage}
\captionof{figure}{Random examples of generated counterfactuals. The first row shows the original image, then the results of applying DeDUCE, JSMA, VLK, and REVISE. The target classes are 9, 9, 3, 6, 9, and 6, respectively. The individual metric scores are reported in table~\ref{tab:testgrid}.}
\label{fig:testgrid}

\begin{minipage}{\linewidth}
\vspace{.5cm}
\begin{center}
\captionof{table}{Metric scores for each of the generated counterfactuals presented in fig.~\ref{fig:testgrid}.}~\\
\label{tab:testgrid}
\begin{tabular}{p{1.3cm}p{1.1cm}p{1.1cm}p{1.1cm}p{1.1cm}p{1.1cm}p{1.1cm}p{1.1cm}}
    \toprule
    & \multicolumn{6}{c}{(original, target)}                   \\
    \cmidrule(r){2-7}
     algorithm & \centering \quad (4,9) & \centering \quad (0,9) & \centering \quad (9,3) & \centering \quad (2,6) & \centering \quad (4,9) & \quad (5,6) & \ metric \\[.05cm]
    \midrule
    \alg{DeDUCE} &
    	\stats{14.01}{5}{3.20} & \stats{21.52}{18}{7.96} & \stats{25.55}{12}{6.75} & 
    	\stats{22.82}{20}{9.19} & \stats{12.23}{3}{2.59} & \stats{17.71}{10}{2.96} & \stats{\emph{AGAN}}{$L_0$}{$L_1$}\\[-.05cm]
    \midrule
    \alg{JSMA} & 
    	\stats{23.31}{29}{11.19}  & \stats{22.32}{15}{10.39} &  \stats{19.86}{19}{8.85} &
    	\stats{25.77}{35}{15.69} & \stats{16.66}{10}{4.73} & \stats{23.01}{6}{2.68} & \stats{\emph{AGAN}}{$L_0$}{$L_1$}\\[-.05cm]
    \midrule
    \alg{VLK} & 
    	\stats{11.77}{60}{10.35} & \stats{56.81}{276}{81.19} & \stats{18.64}{141}{27.28} &
    	\stats{23.06}{177}{48.84} & \stats{18.41}{93}{17.37} & \stats{22.70}{160}{36.72} & \stats{\emph{AGAN}}{$L_0$}{$L_1$}\\[-.05cm]
    \midrule
    \alg{REVISE} & 
    	\stats{15.13}{784}{31.90} & \stats{21.47}{784}{66.08} & \stats{27.41}{692}{57.63} &
    	\stats{N/A}{N/A}{N/A} & \stats{5.49}{784}{17.60} & \stats{25.80}{784}{62.13} & 
    	\stats{\emph{AGAN}}{$L_0$}{$L_1$} \\
    \bottomrule
\end{tabular}
\end{center}
\end{minipage}
\end{figure}
As pointed out above, the GAN-based realism metric is only partly reliable.
This can already be seen from the observation that VAE reconstructions are generally judged to be more realistic than the original image.
Therefore, a qualitative examination of the generated counterfactuals is also warranted, although a conclusive verdict would require a comprehensive human evaluation study.
Individual examples generated by the different algorithms for randomly drawn images and targets are presented in figure~\ref{fig:testgrid}.\footnote{For both the image ids and the target classes, eight integers from 0 to 9 were drawn at random. One pair was drawn twice while for another pair, the target was equal to the label, resulting in six id-target pairs.}
Note that REVISE failed to output a counterfactual in the fourth column.
Column 2 and 4 seem generally difficult, while all approaches arguably find good counterfactuals for column 1 and 5, with the exception of JSMA on the former.
Column 3 and 6 are met with varying success.
Overall, DeDUCE and VLK might be seen to provide the best counterfactuals.
The individual metric scores are presented in table~\ref{tab:testgrid}.
Despite the lack of a user study, it seems clear that the NLL and AGAN scores do not always reflect human judgement.
Because of this and the much higher $L_0$ and $L_1$ scores, the low realism scores of counterfactuals generated by VLK and REVISE clearly do not imply that they are more interpretable than the ones generated by DeDUCE.


\section{Discussion}

As demonstrated in the experiments, DeDUCE is an efficient algorithm performing small and sparse perturbations that often result in realistic counterfactuals.
In particular, it provides counterfactuals that are much more similar to the original image than the other considered approaches.
This allows to give more precise explanations for the model's decision making.
As discussed, DeDUCE is only applicable to classifiers that satisfy sensitivity and smoothness assumptions.
It is sufficient to have a ResNets with some loose spectral normalisation, which might be desired anyway.
Still, this clearly restricts the applicability of DeDUCE.
Overall, DeDUCE could prove to be a viable technique for a considerable number of use cases, namely image classification tasks that require the deployment of large neural networks.

One limitation of this work is the lack of human evaluation studies.
In order to conclusively assess how interpretable the generated counterfactuals are and how helpful the explanations are for debugging, such studies will eventually be necessary.
Another limitation is the lack of demonstrations on more complex datasets.
While MNIST allows a first proof of concept, the algorithm is designed to be scalable to larger datasets and should also be assessed there.

Lastly, future work could also try to improve the algorithm itself.
In particular, it might be possible to use a different latent density model instead of the one used here.
For example, the GMM could be fitted to ambiguous data (using multiple labels and a probabilistic fit) to improve density estimation on such inputs.
This could make it necessary to also train the classification model on ambiguous data.
Another option could be to use confidence-weighting of the datapoints for fitting the class-wise Gaussians.
We leave these ideas as avenues for future research.

\section*{Acknowledgements}

We would like to thank Andreas Kirsch for helpful discussions.
B.H. was supported through a DAAD scholarship.
L.S. and J.M.B. were supported by DeepMind and Cancer Research UK, respectively.
Both L.S. and J.M.B. were also supported by the EPSRC Centre for Doctoral Training in Autonomous Intelligent Machines and Systems (EP/S024050/1).

\newpage
\bibliography{library}

\newpage

\appendix

\section{Effect of different gradient expressions}
\label{app:gradients}

\begin{center}

\captionof{table}{Effect of different objective functions with different hyperparameter settings. Settings with $\lambda$ use a loss function as in equation~(\ref{eq:loss_fct}), whereas settings with $\mu$ use the alternative gradient as in equation~(\ref{eq:grad_sum}). The alternative gradient with no further weighting ($\mu=1$) works best on all metrics. In particular, not including the classification loss at all ($\lambda=0$) performs much worse.}~\\ \label{tab:obj}
\begin{tabular}{c|rrrr}
\toprule
setting & $L_0$ & $L_1$ & \emph{failure} \\
\midrule
$\lambda = 0$ & 27.36 & 13.47 & 0.1\% \\
$\lambda = 1$ & 27.29 & 13.45  & 0.1\% \\
$\lambda = 10$ & 27.16 & 13.40  & 0.1\% \\
$\lambda = 10^2$ & 25.88 & 12.92  & 0\% \\
$\lambda = 10^3$ & 24.58 & 12.38 & 0\% \\
$\lambda = 10^4$ & 25.10 & 12.53 & 0\% \\
$\lambda = 10^5$ & 25.38 & 12.65 & 0\% \\
$\mu = \frac{1}{5}$ & 25.45 & 12.75 & 0\% \\
$\mu = 1$ & \textbf{24.47} & \textbf{12.32} & 0\% \\
$\mu = 5$ & 24.92 & 12.46 & 0.1\% \\
\bottomrule
\end{tabular}
\end{center}

\section{REVISE implementation}
\label{app:revise}

\begin{center}
\includegraphics[width=.9\linewidth]{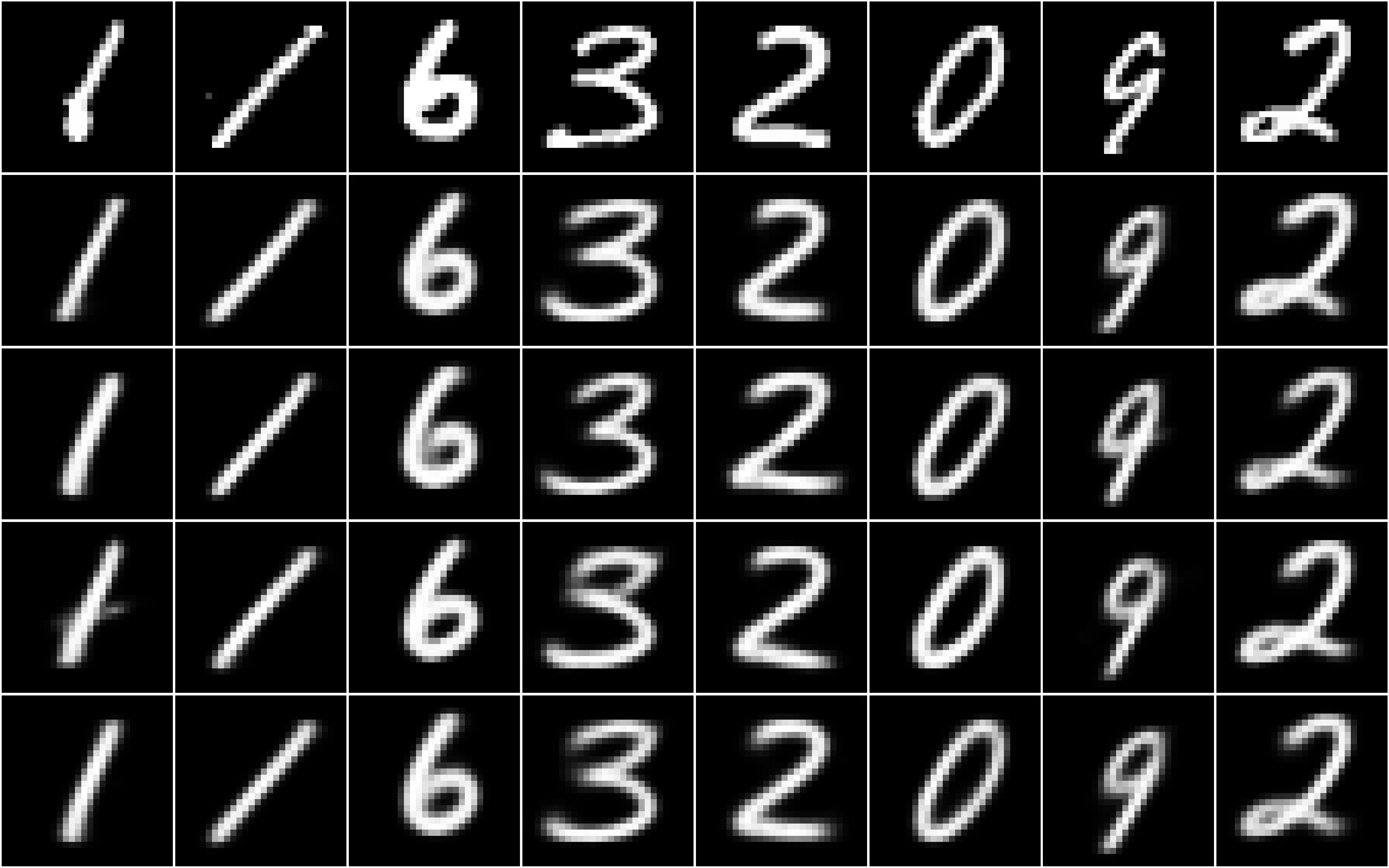}
\captionof{figure}{Some random reconstructions from the VAE used for REVISE. The first row shows randomly selected original MNIST images, followed by three rows of sample reconstructions and the average of 50 reconstructions in the last row.}
\label{fig:revise_recon}
\end{center}

Recall that REVISE takes small steps in the latent space of the VAE, guided by the gradient of the loss function $\ell(f(G(\bz')), t) + \lambda \cdot \| G(\bz') - \bx \|_1$ where $f$ is the classifier, $t$ is the target, $G$ is the generative model, and $\bx$ is the original image.
As in the original paper, we use the cross-entropy loss function for $\ell$.
In addition to the VAE, it is therefore necessary to tune $\lambda$ as well as the gradient step size that we denote by $\delta$.
I perform a grid search on $\lambda \in \{0.1, 1, 10\}$ and $\delta \in \{10^{-3}, 10^{-4}, 10^{-5}\}$, considering both qualitative and quantitative performance.
For the three setting of $\lambda$, we limit the number of iterations to 50,000, 10,000, and 5,000, respectively.\footnote{In all three cases, less than 5\% of the runs terminated in the last 60\% of the iterations, i.e. after step 20,000, 4,000, and 2,000, respectively. This shows that to significantly decrease the failure rate, the iteration limits would need to be raised by more than an order of magnitude, if that helps at all. This is taken as a justification for keeping them at their present values. As a comparison, recall that we limit DeDUCE (and JSMA) to 700 iterations.}
For $\lambda = 10$, the algorithm hardly ever terminates: in all settings for $\delta$, it has a failure rate of over $75\%$.
Comparing all settings on the same image-target pairs would then mean to leave out the vast majority of counterfactuals, so I report quantitative evaluations  only for the other six settings in table~\ref{tab:revise}.
Most notably, for $\lambda = 0.1$, the $L_1$ values are very high; a look at the generated images confirms that these are too far from the original inputs to be useful.
Given $\lambda = 1$, the setting with $\delta =  10^{-5}$ performs clearly the best overall, so we adopt this for the evaluation on the testset.

\begin{center}
\captionof{table}{Performance of REVISE in different settings of $\lambda$ and $\delta$. For $\lambda=0.1$, the counterfactuals are very dissimilar to the original input (see $L_1$) and among settings with $\lambda = 1$, the one with $\delta =  10^{-5}$ performs best overall. Settings with $\lambda = 10$ are not included as the failure rate is above $75\%$.}~\\
\label{tab:revise}
\begin{tabular}{ccrrrr}
\toprule
$\lambda$ & $\delta$ & $L_0$ & $L_1$ & \emph{failure} \\
\midrule
0.1 & $10^{-3}$  & 754.34 & 88.79  & 16.1\% \\
0.1 & $10^{-4}$  & 758.12  & 91.43 & 12\% \\
0.1 & $10^{-5}$ & 764.93 & 84.92 & 11.8\% \\
1 & $10^{-3}$ & 767.75 & 58.66 & 23\% \\
1 & $10^{-4}$ & 771.78 & 54.72 & 26.3\% \\
1 & $10^{-5}$ & 771.89 & 54.11 & 22.2\%  \\
\bottomrule
\end{tabular}
\end{center}

\section{Additional DeDUCE outputs}

\centering
\includegraphics[width=.8\linewidth]{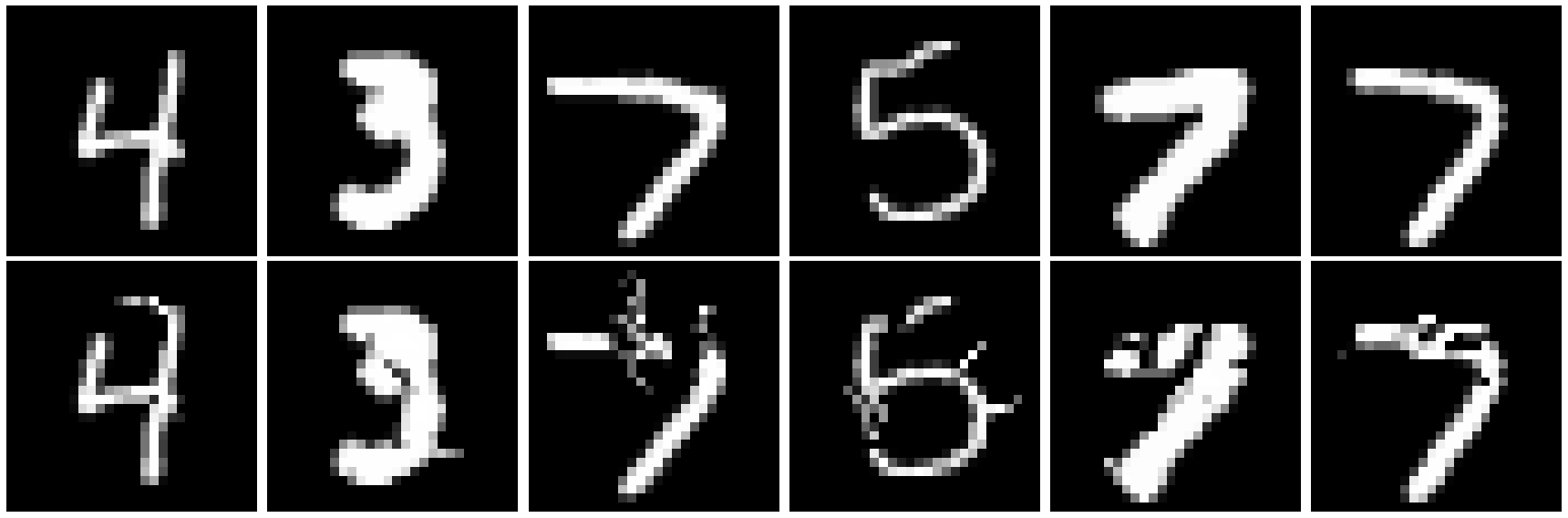}\\[.2cm]
\includegraphics[width=.8\linewidth]{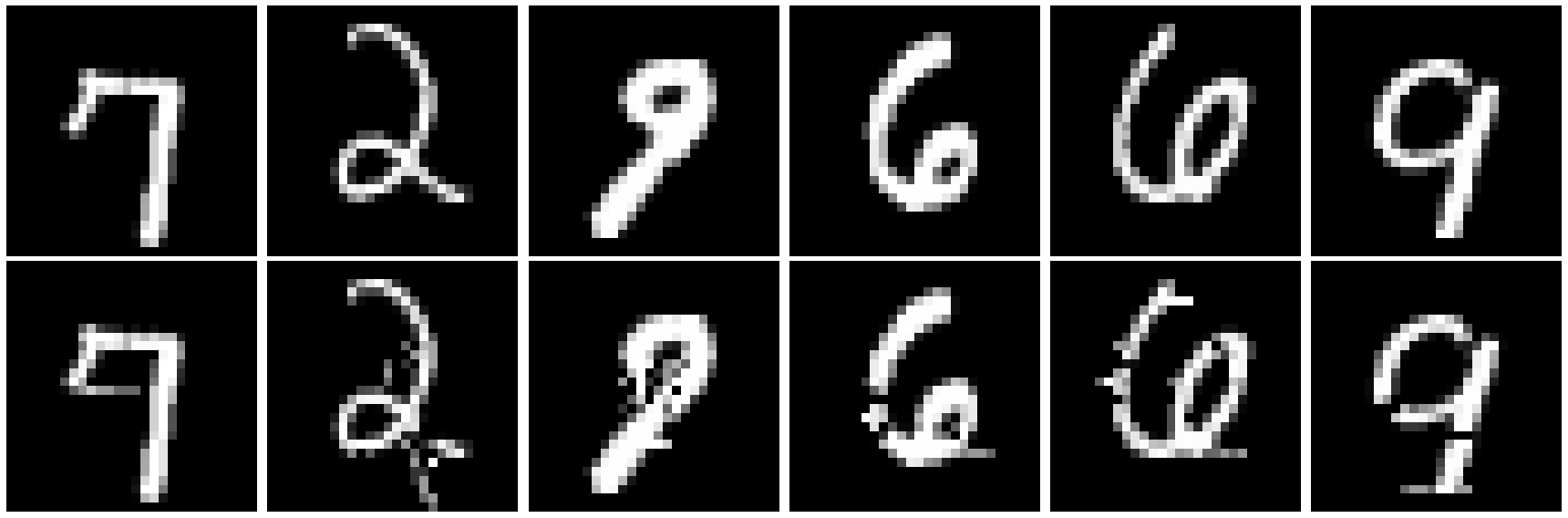}\\[.2cm]
\includegraphics[width=.8\linewidth]{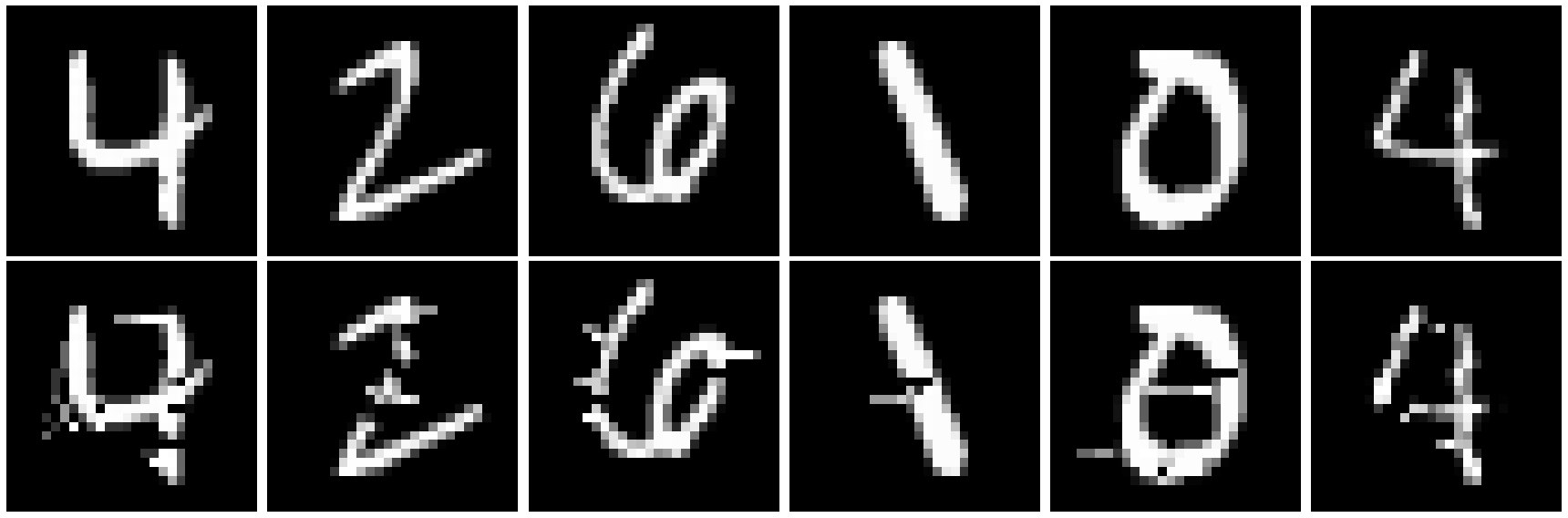}
\captionof{figure}{Additional randomly selected examples produced by DeDUCE. The targets are: \mbox{\textbf{in row two} 9, 1, 4, 4, 1, 9}; \mbox{\textbf{in row four} 9, 9, 0, 1, 2, 3}; \mbox{\textbf{in row six} 0, 5, 5, 2, 9, 1}. In general, the algorithm seems to have the largest difficulties with generating 0s and 1s.}

\end{document}